\documentclass[final]{cvpr}
\usepackage{caption} 
\usepackage{subcaption}
\usepackage{floatrow}

\usepackage{multirow}
\usepackage{times}
\usepackage{epsfig}
\usepackage{graphicx}
\graphicspath{ {images/} }
\usepackage{amsmath}
\usepackage{amssymb}
\DeclareMathOperator*{\argmin}{min}

\newcommand\Mark[1]{\textsuperscript#1}

\usepackage[pagebackref=true,breaklinks=true,colorlinks,bookmarks=false]{hyperref}

\def\cvprPaperID{****} % *** Enter the CVPR Paper ID here
\def\confYear{CVPR 2021}

\begin{document}

%%%%%%%%% TITLE
\title{Beyond Joint Demosaicking and Denoising: An Image Processing Pipeline for a Pixel-bin Image Sensor}

\author{S M A Sharif \Mark{1} , Rizwan Ali Naqvi \Mark{2} \thanks{Corresponding author} , Mithun Biswas \Mark{1} \\
\Mark{1} Rigel-IT, Bangladesh, \Mark{2} Sejong University, South Korea \\
{\tt\small \{sma.sharif.cse,mithun.bishwash.cse\}@ulab.edu.bd, rizwanali@sejong.ac.kr }

}

\maketitle

%%%%%%%%% ABSTRACT
\begin{abstract}
Pixel binning is considered one of the most prominent solutions to tackle the hardware limitation of smartphone cameras. Despite numerous advantages, such an image sensor has to appropriate an artefact-prone non-Bayer colour filter array (CFA) to enable the binning capability. Contrarily, performing essential image signal processing (ISP) tasks like demosaicking and denoising, explicitly with such CFA patterns, makes the reconstruction process notably complicated. In this paper, we tackle the challenges of joint demosaicing and denoising (JDD) on such an image sensor by introducing a novel learning-based method. The proposed method leverages the depth and spatial attention in a deep network. The proposed network is guided by a multi-term objective function, including two novel perceptual losses to produce visually plausible images. On top of that, we stretch the proposed image processing pipeline to comprehensively reconstruct and enhance the images captured with a smartphone camera, which uses pixel binning techniques. The experimental results illustrate that the proposed method can outperform the existing methods by a noticeable margin in qualitative and quantitative comparisons. Code available: \url{https://github.com/sharif-apu/BJDD_CVPR21}.
\end{abstract}
%%%%%%%%%%%%%%%%%%%%%%%%%%%%%%%%%%%%%%%%%%
\vspace{-1.\baselineskip}
%\vspace*{-12px}
%%%%%%%%% BODY TEXT
\section{Introduction}
Smartphone cameras have illustrated a significant altitude in the recent past. However, the compact nature of mobile devices noticeably impacts the image quality compared to their DSLR counterparts \cite{ignatov2017dslr}. Also, such inevitable hardware limitations, holding back the original equipment manufacturers (OEMs) to achieve a substantial jump in the dimension of the image sensors. In contrast, the presence of a bigger sensor in any camera hardware can drastically improve the photography experience, even in stochastic lighting conditions \cite{liu2007high}. Consequently, numerous OEMs have exploited pixel enlarging techniques known as pixel binning in their compact devices to deliver visually admissible images \cite{barna2013method,yoo2015low}.      

In general, pixel binning aims to combine the homogenous neighbour pixels to form a larger pixel \cite{agranov2017pixel}. Therefore, the device can exploit a larger sensor dimension outwardly incorporating an actual bigger sensor. Apart from leveraging a bigger sensor size in challenging lighting conditions, such image sensor design also has substantial advantages. Among them, capture high-resolution contents, producing a natural bokeh effect, enable digital zoom by cropping an image, etc., are noteworthy. This study denotes such image sensors as a pixel-bin image sensor.

\begin{figure}[!ht]
    \captionsetup[subfigure]{labelformat=empty}
     \centering
     \begin{subfigure}[b]{3.1cm}
         \centering
         \includegraphics[width=3.1cm]{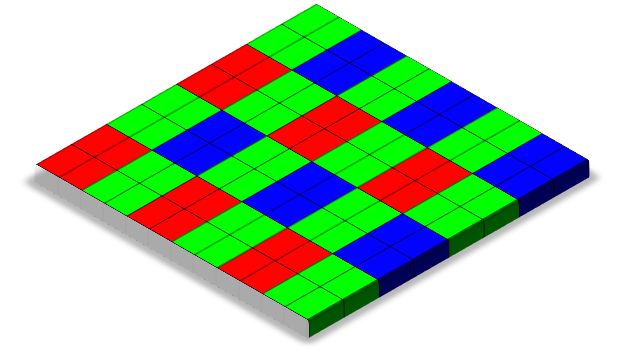}
         \caption{Quad Bayer CFA}
         \label{QBCFA}
     \end{subfigure}
     %\hfill
     \begin{subfigure}[b]{3.1cm}
         \centering
         \includegraphics[width=3.1cm]{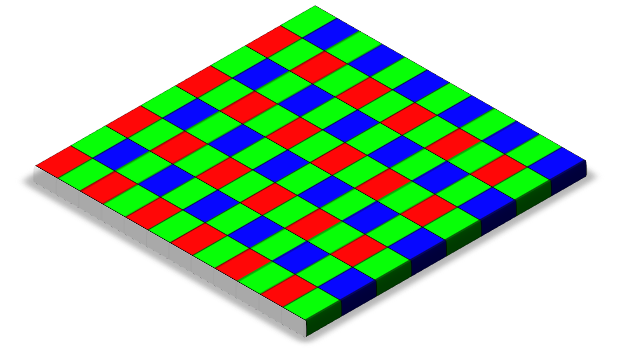}
         \caption{Bayer CFA}
         \label{BCFA}
     \end{subfigure}
     \hfill
        \caption{Commonly used CFA patterns of pixel-bin image sensors.}
        \label{CFA}
\end{figure}

Despite the widespread usage in recent smartphones, including Oneplus Nord, Galaxy S20 FE, Xiaomi Redmi Note 8 Pro, Vivo X30 Pro, etc., reconstructing RGB images from a pixel-bin image sensor is notably challenging \cite{kim2019deep}. Expressly, the pixel binning techniques have to employ a non-Bayer CFA \cite{lahav2010color,kim2019deep} along with a traditional Bayer CFA \cite{bayer1976color} over the image sensors to leverage the binning capability. Fig. \ref{CFA} depicts the most commonly used CFA patterns combination used in recent camera sensors. Regrettably, the non-Bayer CFA  (i.e., Quad Bayer CFA \cite{kim2019high}) has to appropriate in pixel-bin image sensors is notoriously vulnerable to produce visually disturbing artefacts while reconstructing images from the given CFA pattern \cite{kim2019deep}. Hence, combining fundamental low-level ISP tasks like denoising and demosaicking on an artefact-prone CFA make the reconstruction process profoundly complicated.

\begin{figure*}[!ht]
    \captionsetup[subfigure]{labelformat=empty,justification=centering}
     \centering
     \begin{subfigure}[b]{2.42cm}
         \centering
         \includegraphics[width=\textwidth]{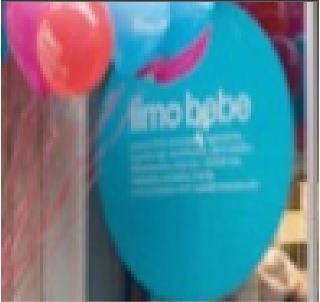}
         \caption{Reference\\PSNR:$\infty$}
         %\label{QBCFA}
     \end{subfigure}
     \hfill
     \begin{subfigure}[b]{2.42cm}
         \centering
         \includegraphics[width=\textwidth]{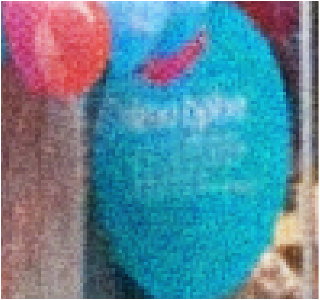}
         \caption{Deepjoint \cite{gharbi2016deep}\\PSNR:28.66 dB}
         %\label{BCFA}
     \end{subfigure}
     \hfill
     \begin{subfigure}[b]{2.42cm}
         \centering
         \includegraphics[width=\textwidth]{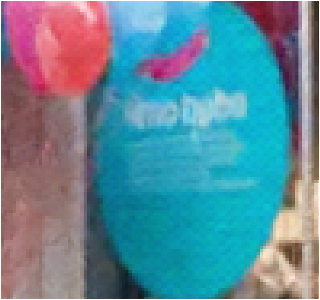}
         \caption{Kokkinos \cite{kokkinos2018deep}\\PSNR:31.02 dB}
         %\label{QBCFA}
     \end{subfigure}
     \hfill
     \begin{subfigure}[b]{2.42cm}
         \centering
         \includegraphics[width=\textwidth]{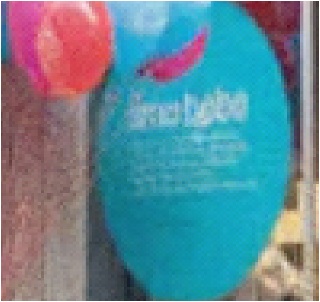}
         \caption{Dong \cite{dong2018joint}\\PSNR:30.07 dB}
         %\label{BCFA}
     \end{subfigure}
     \hfill
     \begin{subfigure}[b]{2.42cm}
         \centering
         \includegraphics[width=\textwidth]{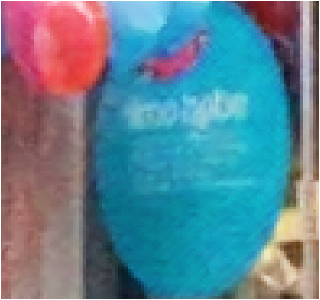}
         \caption{DeepISP \cite{schwartz2018deepisp}\\PSNR:31.65 dB}
         %\label{BCFA}
     \end{subfigure}
     \hfill
     \begin{subfigure}[b]{2.42cm}
         \centering
         \includegraphics[width=\textwidth]{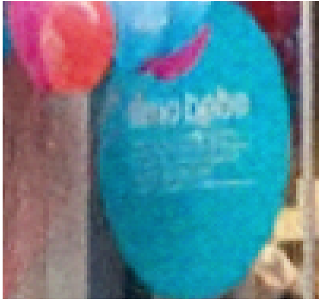}
         \caption{DPN \cite{kim2019deep}\\PSNR:32.59 dB}
         %\label{BCFA}
     \end{subfigure}
     \hfill
     \begin{subfigure}[b]{2.42cm}
         \centering
         \includegraphics[width=\textwidth]{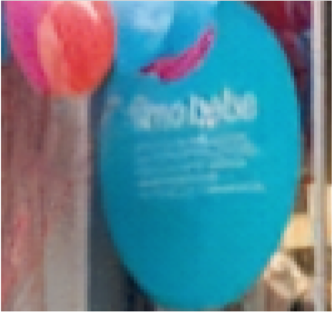}
         \caption{\textbf{Ours\\PSNR:34.81 dB}}
         %\label{BCFA}
     \end{subfigure}
     \hfill
      \vspace{-1\baselineskip}
        \caption{Example of Joint demosaicing and denoising on Quad Bayer CFA.}
        \label{intro}
\end{figure*}

Contrarily, the learning-based methods have illustrated distinguished progression in performing image reconstruction tasks. Also, they have demonstrated substantial advantages of combining low-level tasks such as demosaicing along with denoising \cite{gharbi2016deep,kokkinos2018deep,liu2020joint,dong2018joint}. Most notably, some of the recent convolutional neural network (CNN) based methods \cite{schwartz2018deepisp,ignatov2020replacing} attempt to mimic complicated mobile ISP and substantiate significant improvement in perceptual quality over traditional methods. Such computational photography advancements inspired this study to tackle the challenging JDD of a  pixel-bin image sensor and go beyond. 

This study introduces a novel learning-based method to perform JDD in commonly used CFA patterns (i.e., Quad Bayer CFA \cite{kim2019high}, and Bayer CFA \cite{bayer1976color}) of pixel-bin image sensors. The proposed method leverage spatial and depth-wise feature correlation \cite{woo2018cbam, hu2018squeeze} in a deep architecture to reduce visual artefacts. We have denoted the proposed deep as a pixel-bin image processing network (PIPNet) in the rest of the paper. Apart from that, we introduced a multi-term guidance function, including two novel perceptual losses to guide the proposed PIPNet for enhancing the perceptual quality of reconstructed images. Fig. \ref{intro} illustrates an example of the proposed method's JDD performance on a non-Bayer CFA. The feasibility of the proposed method has extensively studied with diverse data samples from different colour spaces. Later, we stretched our proposed pipeline to reconstruct and enhance the images of actual pixel-bin image sensors.  
%\vspace{-1.\baselineskip}

The contribution of this study has summarized below:
 \begin{itemize}
 \vspace*{-1px}
     \item Proposes a learning-based method, which aims to tackle the challenging JDD on a pixel-bin image sensor. 
     \item Proposes a deep network that exploits depth-spatial feature attentions and is guided by a multi-term objective function, including two novel perceptual losses. 
     \item Stretches the proposed method to study the feasibility of enhancing perceptual image quality along with JDD on actual hardware.
 \end{itemize}
%%%%%%%%%%%%%%%%%%%%%%%%%%%%%%%%%%%%%%%%%%

%%%%%%%%%%%%%%%%%%%%%%%%%%%%%%%%%%%%%%%%%%
\section{Related work}
This section briefly reviews the works that are related to the proposed method.
%%%%%%%%%%%%%%%%%%%%%%%%%%%%%%%%%%%%%%%%%%

\textbf{Joint demosacing and denoising.}
Image demosaicing is considered a low-level ISP task, aiming to reconstruct RGB images from a given CFA pattern. However, in practical application, the image sensors' data are contaminated with noises, which directly costs the demosaicking process by deteriorating final reconstruction results \cite{liu2020joint}. Therefore, the recent works emphasize performing demosaicing and denoising jointly rather than traditional sequential approaches. 

In general, JDD methods are clustered into two major categories: optimization-based methods \cite{hirakawa2006joint, tan2017joint} and learning-based methods \cite{gharbi2016deep, dong2018joint,kokkinos2018deep}.  However, the later approach illustrates substantial momentum over their classical counterparts, particularly in reconstruction quality. In recent work, numerous novel CNN-based methods have been introduced to perform the JDD. For example, \cite{gharbi2016deep} trained and a deep network with millions of images to achieve state-of-the-art results. Similarly, \cite{kokkinos2018deep} fuse the majorization-minimization techniques into a residual denoising network, \cite{dong2018joint} proposed a generative adversarial network (GAN) along with perceptual optimization to perform JDD. Also, \cite{liu2020joint} proposed a deep-learning-based method supervised by density-map and green channel guidance. Apart from these supervised approaches, \cite{ehret2019joint} attempts to solve  JDD with unsupervised learning on burst images.
%%%%%%%%%%%%%%%%%%%%%%%%%%%%%%%%%%%%%%%%%%

\textbf{Image enhancement.}
Image enhancement works mostly aim to improve the perceptual image quality by incorporating colour correction, sharpness boosting, denoising, white balancing, etc. Among the recent works, \cite{fu2016fusion,yuan2012automatic} proposed learning-based solutions for automatic global luminance and gamma adjustment. Similarly, \cite{lee2016automatic} offered deep-learning solutions for colour and tone correction, and \cite{yuan2012automatic} presented a CNN model to image contrast enhancement. However, the most comprehensive image enhancement approach was introduced by \cite{ignatov2017dslr}, where the author enhanced downgraded smartphone images according to superior-quality photos obtained with a high-end camera system.
%%%%%%%%%%%%%%%%%%%%%%%%%%%%%%%%%%%%%%%%%%

\textbf{Learning ISP.}
A typical camera ISP pipeline exploits numerous image processing blocks to reconstruct an sRGB image from the sensor's raw data. A few novel methods have recently attempted to replace such complex ISPs by learning from the convex set of data samples. In \cite{schwartz2018deepisp}, the authors proposed a CNN model to suppress image noises and exposure correction of images captured with a smartphone camera.  Likewise, \cite{ignatov2020replacing} proposed a  deep model incorporating extensive global feature manipulation to replace the entire ISP of the Huwaei P20 smartphone. In another recent work, \cite{liang2019cameranet} proposed a two-stage deep network to replicate camera ISP. 
%%%%%%%%%%%%%%%%%%%%%%%%%%%%%%%%%%%%%%%%%%

\textbf{Quad Bayer Reconstruction.} 
Reconstructing RGB images from a Quad Bayer CFA is considerably challenging. In \cite{kim2019deep} has addressed this challenging task by proposing a duplex pyramid network. It worth noting, none of the existing methods (including \cite{kim2019deep}) specialized for our target applications. However, their respective domains' success inspired this work to develop an image processing pipeline for a pixel-bin image sensor, which can perform JDD and go beyond.
%%%%%%%%%%%%%%%%%%%%%%%%%%%%%%%%%%%%%%%%%%

%%%%%%%%%%%%%%%%%%%%%%%%%%%%%%%%%%%%%%%%%%
\section{Method}
This section details the network design, a multi-term objective function, and implementation strategies.

\begin{figure*}[!htb]
\centering
\includegraphics[width=13cm,keepaspectratio]{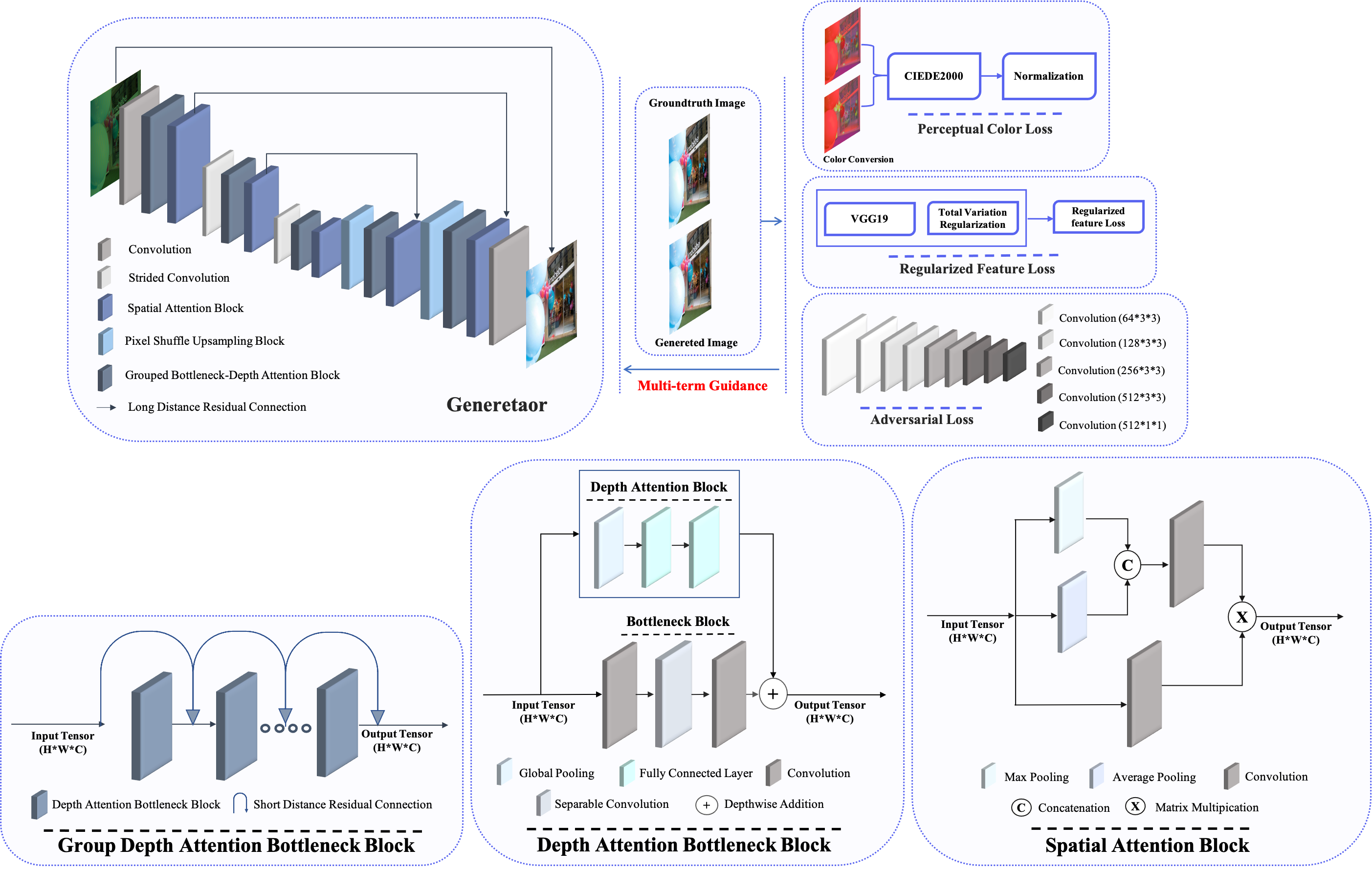}
\caption{  Overview of the proposed method, including network architecture and submodules. }
\label{network}
\end{figure*}

\subsection{Network design}
Fig. \ref{network} depicts the proposed method's overview, including the novel PIPNet architecture. Here, the proposed network exploits feature correlation, also known as attention mechanism \cite{hu2018squeeze,woo2018cbam,dai2019second}, through the novel components in U-Net \cite{ronneberger2015u} like architecture to mitigate visual artefacts. Overall, the method aims to map a mosaic input ($I_M$) as $\mathrm{G}: \mathbf{I_M} \to  \mathbf{I_R}$. Where the mapping function ($\mathrm{F}$) learns to reconstruct an RGB image ($\mathbf{I_R}$)  as $\mathbf{I_R} \in [0,1]^{H \times W \times 3}$. $H$ and $W$ represent the height and width of the input and output images. 

\textbf{Group depth attention bottleneck block.}
The novel group depth attention bottleneck (GDAB) block allowed the proposed network to go deeper by leveraging depth attention \cite{hu2018squeeze}. The GDAB block comprises of $\mathbf{m} \in \mathbb{Z}$ number of depth attention bottleneck (DAB) blocks. Where the DABs are stacked consecutively and connected with short distance residual connection; thus, the network can converge with informative features \cite{dai2019second}. For any $g$-th  member of a GDAB block can be represented as:

\begin{equation}
\mathbf{F}_g = \mathbf{W}_g\mathbf{F}_{g-1} + H_g(\mathbf{F}_{g-1})
\end{equation}
Here, $\mathbf{W}_g$, $\mathbf{F}_{g-1}$, and $\mathbf{F}_{g}$ represent the corresponding weight matrics, input, and output features. $H_g(\cdot)$ denotes the function of group members (i.e., DAB).

\textbf{Depth attention bottleneck block.}
The proposed DAB incorporates a depth attention block along with a bottleneck block. For a given input $\mathbf{X}$, the $m$-th DAB block aims to output the feature map $\mathbf{X}^{\prime}$ as:  
\begin{equation}
\mathbf{X}_m^{\prime} = B_m(\mathbf{X}) + {D}_m(\mathbf{X})
\label{dabEQ}
\end{equation}
In Eq. \ref{dabEQ}, $B(\cdot)$ presents the bottleneck block function, which has been inspired by the well-known MobileNetV2 \cite{sandler2018mobilenetv2}. The main motive of utilizing the bottleneck  block is to control the trainable parameters with satisfactory performance. Typically, pixel-bin image sensors are exclusively designed for mobile devices. Therefore, we stress to reduce the trainable parameters as much as possible. Apart from the bottleneck block, DAB also incorporates a depth attention block, which has denoted as $D(\cdot)$ in Eq. 2. It is worth noting, this study proposes to adding the feature map of the depth attention block along with the bottleneck block to leverage the long-distance depth-wise attention \cite{hu2018squeeze,dai2019second}.  Here, depth-wise squeezed descriptor $\mathbf{Z} \in \mathbb{R}^C$ has been obtained by shrinking $\hat{\mathbf{X}} = [\mathbf{x}_1 , \dots, \mathbf{x}_c]$  as follows:

%As illustrated in Fig. 2, let Yˆ = [y1 , · · · , yC ], the channel-wise statistics z ∈ RC×1 can be obtained by shrinking Yˆ. Then the c-th dimension of z is computed as
 
\begin{equation}
%\mathbf{Z}_{c} = \frac{1}{H \times W}\sum_{i=1}^{H}\sum_{j=1}^{W}\mathbf{X}_{c}(i, j) 
\mathbf{Z}_{c} = \mathbf{A}_{GP}(\mathbf{x}_c) = \frac{1}{C}\sum_{i}^{C}\mathbf{x}_{c}(i) 
\end{equation}
Here, $\mathbf{A}_{GP}(_c)$ presents the global average pooling, spatial dimension, and feature map.

Additionally, an aggregated global dependencies have pursued by applying a gating mechanism  as follows:
 
\begin{equation}
\mathbf{W} = \tau(\mathbf{W_{S}}(\delta(\mathbf{W_{R}}(\mathbf{Z}))))
\end{equation}
Here, $\tau$ and $\delta$ represent the sigmoid and ReLU activation functions, which have applied after $\mathbf{W_{S}}(\cdot)$ and $\mathbf{W_{R}}(\cdot)$ convolutional operations, which intended to set depth dimension of features to $\mathbf{C/r}$ and $\mathbf{C}$.

The final output of the depth attention block has obtained by applying a depth-wise attention map with a rescaling factor \cite{dai2019second} described as follows:
 
\begin{equation}
\mathbf{D_{c}} = \mathbf{W_{c}}\cdot\mathbf{S_{c}}
\end{equation}
Here, $\mathbf{W_{c}}$ and $\mathbf{S_{c}}$ represent the feature map and scaling factor. 

\textbf{Spatial attention block.}
The spatial attention block of the proposed method has been inspired by recent convolutional spatial modules \cite{woo2018cbam,chen2017sca}. It aims to realize the spatial feature correlation from a given feature map $\mathbf{X}$ as follows:
\begin{equation}
%\mathbf{F} = \tau(\mathbf{F_S}([\mathbf{Z_{A}(X)};\mathbf{Z_M(X)}])) \times \mathbf{F_S}(\mathbf{X})
\mathbf{F} = \tau(\mathbf{F_S}([\mathbf{Z_{A}(X)};\mathbf{Z_M(X)}])
\end{equation}
Here, $\mathbf{F}(\cdot)$ and $\tau$ represent the convolution operation and sigmoid activation. Additionally, $\mathbf{Z_{A}}$ and $\mathbf{Z_{M}}$ present the average pooling and max pooling, which generates two 2D feature map as $\mathbf{X_{A}} \in \mathbb{R}^{1 \times H \times W}$  and $\mathbf{X_{M}} \in \mathbb{R}^{1 \times H \times W}$. 

 \textbf{Transition Layer}. The proposed network traverses different features depth to exploit the UNet like structure using upscaling or downscaling operations. The downsampling operation has obtained on an input feature map $\mathbf{X}_{0}$ as follows:
 
\begin{equation}
\mathbf{F}_{\downarrow} = H_{\downarrow} (\mathbf{X}_{0})
\end{equation}
Here, $H_{\downarrow}(\cdot)$ represents a stride convolutional operation.

Inversely, the upscaling on an input feature map $\mathbf{X}_{0}$ has achieved as:
 
\begin{equation}
\mathbf{F}_{\uparrow} = H_{\uparrow} (\mathbf{X}_{0})
\end{equation}
Here, $H_{\uparrow}(\cdot)$ represents the pixel shuffle convolution operation followed by the PReLU function, which intends to avoid checkerboard artefacts \cite{aitken2017checkerboard}.

\textbf{Conditional Discriminator.}
The proposed PIPNet has appropriated the concept of adversarial guidance and adopted a well-established conditional Generative Adversarial Network (cGAN) \cite{mirza2014conditional}. The objective of the cGAN discriminator consists of stacked convolutional operations and set to maximize as: $\mathbb{E}_{\mathbf{X,Y}} \big[\log D\big({\mathbf{X,Y}}\big) \big]$.
 
\subsection{Objective function}
The proposed network $\mathrm{G}$ parameterized with weights $\mathbf{W}$, aims to minimize the training loss by appropriating the given $P$ pairs of training images $\{ \mathbf{I_M}^t, \mathbf{I_G}^t \}_{t=1}^P$ as follows:

\begin{equation}
 W^\ast = \arg{\argmin_W}\frac{1}{P}\sum_{t=1}^{P}\mathcal{L}_{\mathit{T}}(\mathrm{G}(\mathbf{I_M}^t), \mathbf{I_G}^t)
 \label{fLoss}
\end{equation}
Here, $\mathcal{L}_{\mathit{T}}$ denotes the proposed multi-term objective function, which aims to improve the perceptual quality (i.e., details, texture, colour, etc.) while reconstructing an image.

\textbf{Reconstruction loss.}
L1-norm is known to be useful for generating sharper images \cite{zhao2016loss, schwartz2018deepisp}. Therefore, an L1-norm has adopted to calculate pixel-wise reconstruction error as follows:

\begin{equation}
 \mathcal{L}_{\mathit{R}} = \parallel \mathbf{I_G} - \mathbf{I_R} \parallel_1
\end{equation}
Here, $\mathbf{I_G}$ and $\mathbf{I_R}$ present the ground truth image and output of $\mathrm{G}(\mathbf{I_M})$ respectively.

\textbf{Regularized feature loss (RFL).}:
VGG-19 feature-based loss functions aim to improve a reconstructed image's perceptual quality by encouraging it to have identical feature representation like the reference images \cite{ignatov2017dslr, mechrez2018contextual, wang2018esrgan}. Typically, such activation-map loss functions represented as follows:

\begin{equation}
\label{RFL}
 \mathcal{L}_{\mathit{FL}} = \lambda_P \times \mathcal{L}_{\mathit{VGG}}
\end{equation}
Where $\mathcal{L}_{\mathit{VGG}}$ can be extended as follows:
\begin{equation}
 \mathcal{L}_{\mathit{VGG}} =
 \frac{1}{H_j W_j  C_j}\parallel \psi_t(\mathbf{I_G}) - \psi_t(\mathbf{I_R}) \parallel_1
\end{equation}
Here, $\psi$ and $j$ denote the pre-trained VGG network and its $j^{th}$ layer.

It is worth noting, in Eq. \ref{RFL}, $\lambda_P$ denotes the regulator of a feature loss. However, in most cases, the regulator's value has to set emphatically, and without proper tuning, it can deteriorate the reconstruction process \cite{wang2018esrgan}. To address this limitation, we replaced $\lambda_P$ with a total variation regularization \cite{strong2003edge}, which can be presented as follows:

\begin{equation}
 \lambda_R = \frac{1}{H_j  W_j  C_j}\parallel \Delta{O_v} \parallel + \parallel \Delta{O_h} \parallel
\end{equation}
Here,$\parallel \Delta{O_v} \parallel$ and $\parallel \Delta{O_h} \parallel$ present the gradients' summation
in the vertical and horizontal directions calculated over a training pair. The regularized form of Eq. \ref{RFL} can be written as:

\begin{equation}
 \mathcal{L}_{\mathit{RFL}} =  \lambda_R \times \mathcal{L}_{\mathit{VGG}}
\end{equation}

\textbf{Perceptual colour loss (PCL).} Due to the smaller aperture and sensor size, most smartphone cameras are prone to illustrate colour inconsistency in numerous instances \cite{ignatov2017dslr}. We developed a CIEDE2000 \cite{luo2001development} based on the perceptual colour loss to address this limitation, which intends to measure the colour difference between two images in euclidean space. Subsequently, the newly developed loss function encourages the proposed network to generate a similar colour as the reference image. The proposed perceptual colour loss can be represented as follows:

\begin{equation}
 \mathcal{L}_{\mathit{PCL}} = \Delta{E} \Big( \mathbf{I_G}, \mathbf{I_R} \Big)
\end{equation}
Here, $\Delta{E}$ represents the CIEDE2000 colour difference \cite{luo2001development}.

\textbf{Adversarial loss.}
Adversarial guidance is known to be capable of recovering texture and natural colours while reconstructing images. Therefore, we encouraged our model to employ a cGAN based cross-entropy loss as follows: 

\begin{equation}
 \mathcal{L}_{\mathit{G}}= - \sum_{t} \log D(\mathbf{I_G}, \mathbf{I_R})
\end{equation}
Here, $D$ denotes the conditional discriminator, which aims to perform as a global critic.

\textbf{Total loss.}
The final multi-term objective function ($\mathcal{L}_{\mathit{T}}$) has calculated as follows:
 
\begin{equation}
 \mathcal{L}_{\mathit{T}}= \mathcal{L}_{\mathit{R}} + \mathcal{L}_{\mathit{RFL}} + \mathcal{L}_{\mathit{PCL}} +  \lambda_{G}.\mathcal{L}_{\mathit{G}}
\end{equation}
Here, $\lambda_{G}$ presents adversarial regulators and set as $\lambda_{G}$ = 1e-4. 

\subsection{Implementation details}

The generator of proposed PIPNet traverses between different feature depth to leverage the UNet like structure as $d = (64,126,256)$, where the GDAB blocks of the proposed network comprise $\mathbf{m}=3$  number of DAB block (also refer as group density in a later section). Every convolution operation in the bottleneck block of a DAB block incorporates $1\times1$ convolutional and a $3\times3$ separable convolution, where each layer is activated with a LeakyReLU function. Additionally, the spatial attention block, downsampling block, and discriminator utilize $3\times3$ convolutional operation. A swish function has activated the convolution operations of the discriminator. Also, every $(2n-1)^{th}$ layer of the discriminator increases the feature depth and reduces the spatial dimension by 2.
%%%%%%%%%%%%%%%%%%%%%%%%%%%%%%%%%%%%%%%%%%

%%%%%%%%%%%%%%%%%%%%%%%%%%%%%%%%%%%%%%%%%%
\section{Experiments}
\label{experiment}
The performance of the proposed method has been studied extensively with sophisticated experiments. This section details the experiment results and comparison for JDD.

%%%%%%%%%%%%%%%%%%%%%%%%%%%%%%%%%%%%%%%%%%
\subsection{Setup}
To learn JDD for a pixel-bin image sensor, we extracted 741,968 non-overlapping image patches of dimension $128 \times 128$ from DIV2K \cite{agustsson2017ntire} and Flickr2K \cite{timofte2017ntire} datasets. The image patches are sampled according to the CFA patterns and contaminated with a random noise factor of $\mathcal{N}(\mathbf{I_G}|\sigma)$. Here, $\sigma$ represents the standard deviation of a Gaussian distribution, which is generated by $\mathcal{N}(\cdot)$ over a clean image $\mathbf{I_G}$. It has presumed that the JDD has performed in sRGB colour space before colour correction, tone mapping, and white balancing. The model has implemented in the PyTorch \cite{pytorch} framework and optimized with an Adam optimizer \cite{kingma2014adam} as $\beta_1 = 0.9$, $\beta_2 = 0.99$, and learning rate = $1e-4$. The model trained for $10 \sim 15$ epoch depending on the CFA pattern with a constant batch size of 12.  The training process accelerated using an Nvidia Geforce GTX 1060 (6GB) graphical processing unit (GPU).

%%%%%%%%%%%%%%%%%%%%%%%%%%%%%%%%%%%%%%%%%%
\subsection{Joint demosaicing and denoising}
We conducted an extensive comparison with the benchmark dataset for the evaluation purpose, including BSD100 \cite{MartinFTM01}, McM \cite{wu2011single}, Urban100 \cite{cordts2016cityscapes}, Kodak \cite{yanagawa2008kodak}, WED \cite{ma2016waterloo}, and MSR demosaicing dataset \cite{khashabi2014joint}. We used only linRGB images from the MSR demosaicing dataset to verify the proposed method's feasibility in different colour spaces (i.e., sRGB and linRGB). Therefore, we denoted the MSR demosaicing dataset as linRGB in the rest of the paper. Apart from that, four CNN-based JDD methods (Deepjoint \cite{gharbi2016deep}, Kokkinos \cite{kokkinos2018deep}, Dong \cite{dong2018joint},  DeepISP \cite{schwartz2018deepisp}) and a specialized Quad Bayer reconstruction method (DPN \cite{kim2019deep}) have studied for comparison. Each compared method's performance cross-validated with three different noise levels $\sigma = (5,15,25$) and summarized with the following evaluation metrics: PSNR, SSIM, and DeltaE2000. 

%%%%%%%%%%%%%%%%%%%%%%%%%%%%%%%%%%%%%%%%%%
\subsubsection{Quad Bayer CFA}
\begin{table*}[htb]
\centering
\scalebox{.65}{\begin{tabular}{llllllll}
\hline
\multirow{2}{*}{\textbf{Method}} & \multirow{2}{*}{\textbf{$\sigma$}} & \textbf{BSD100}            & \textbf{WED}               & \textbf{Urban100}          & \textbf{McM}               & \textbf{Kodak}             & \textbf{linRGB}            \\ \cline{3-8}
                                 &                              & \textbf{PSNR/SSIM/DeltaE}  & \textbf{PSNR/SSIM/DeltaE}  & \textbf{PSNR/SSIM/DeltaE}  & \textbf{PSNR/SSIM/DeltaE}  & \textbf{PSNR/SSIM/DeltaE}  & \textbf{PSNR/SSIM/DeltaE}  \\ \hline
Deepjoint \cite{gharbi2016deep}                          & \multirow{6}{*}{5}              & 34.69/0.9474/2.45          & 30.99/0.9115/3.30           & 31.04/0.9272/3.43          & 31.16/0.8889/3.21          & 32.39/0.9310/3.01          & 40.03/0.9694/1.59          \\
Kokkinos \cite{kokkinos2018deep}                             &                                 & 34.54/0.9662/2.25          & 32.18/0.9428/2.77          & 32.16/0.9501/3.13          & 32.84/0.9280/2.49          & 33.05/0.9532/2.74          & 38.46/0.9670/1.45          \\
Dong \cite{dong2018joint}                             &                                 & 33.79/0.9604/2.39          & 31.22/0.9312/3.03          & 31.48/0.9395/3.24          & 31.80/0.9117/2.81          & 32.72/0.9494/2.85          & 38.40/0.9477/1.75          \\
DeepISP \cite{schwartz2018deepisp}                              &                                 & 36.78/0.9711/1.94          & 32.25/0.9392/2.93          & 32.52/0.9494/3.10          & 32.32/0.9143/2.89          & 34.25/0.9547/2.45          & 42.02/0.9819/1.27          \\
DPN \cite{kim2019deep}                           &                                 & 37.71/0.9714/1.81          & 33.60/0.9490/2.50           & 33.73/0.9560/2.78          & 34.37/0.9354/2.26          & 35.73/0.9597/2.22          & 42.08/0.9767/1.16          \\
\textbf{PIPNet}                    &                                 & \textbf{39.43/0.9803/1.42} & \textbf{35.08/0.9609/2.08} & \textbf{35.66/0.9670/2.25} & \textbf{35.55/0.9470/1.90} & \textbf{37.21/0.9699/1.79} & \textbf{44.14/0.9827/0.94} \\ \hline
Deepjoint \cite{gharbi2016deep}                          & \multirow{6}{*}{15}             & 32.68/0.8883/2.81          & 29.97/0.8678/3.61          & 30.02/0.8819/3.72          & 30.19/0.8432/3.51          & 31.02/0.8655/3.35        & 36.94/0.9100/1.88          \\
Kokkinos \cite{kokkinos2018deep}                             &                                 & 33.57/0.9396/2.56          & 31.38/0.9183/3.07          & 31.40/0.9262/3.38          & 32.03/0.9026/2.79          & 32.17/0.9208/3.04          & 37.72/0.9453/1.75          \\
Dong \cite{dong2018joint}                            &                                 & 32.49/0.9280/2.67          & 30.27/0.9004/3.32          & 30.52/0.9110/3.50           & 30.80/0.8755/3.08          & 31.58/0.9128/3.13          & 35.66/0.9061/2.05          \\
DeepISP \cite{schwartz2018deepisp}                              &                                 & 34.73/0.9443/2.30           & 31.31/0.9174/3.20           & 31.44/0.9276/3.37          & 31.47/0.8909/3.13          & 32.74/0.9226/2.81          & 39.44/0.9608/1.59          \\
DPN \cite{kim2019deep}                             &                                 & 35.08/0.9451/2.17          & 32.27/0.9265/2.80           & 32.22/0.9340/3.07          & 33.16/0.913/2.53           & 33.60/0.9274/2.57          & 39.43/0.9613/1.44          \\
\textbf{PIPNet}                    &                                 & \textbf{36.68/0.9586/1.79} & \textbf{33.55/0.9416/2.41} & \textbf{33.85/0.9481/2.56} & \textbf{34.21/0.9277/2.18} & \textbf{34.89/0.9422/2.14} & \textbf{41.86/0.9728/1.18} \\ \hline
Deepjoint \cite{gharbi2016deep}                          & \multirow{6}{*}{25}             & 30.15/0.8006/3.51          & 28.42/0.8008/4.16          & 28.44/0.8140/4.28          & 28.69/0.7745/4.03          & 29.00/0.7724/4.02          & 33.73/0.8246/2.51          \\
Kokkinos \cite{kokkinos2018deep}                             &                                 & 31.99/0.9070/2.95           & 30.12/0.8868/3.45          & 30.09/0.8961/3.74          & 30.74/0.8697/3.17          & 30.74/0.8827/3.43          & 36.03/0.9152/2.08          \\
Dong \cite{dong2018joint}                            &                                 & 31.49/0.8870/3.00          & 29.50/0.8607/3.60            & 29.72/0.8755/3.80          & 29.67/0.8205/3.38          & 30.76/0.8685/3.43          & 31.74/0.8419/2.65          \\
DeepISP \cite{schwartz2018deepisp}                              &                                 & 32.85/0.9111/2.79          & 30.17/0.8885/3.60           & 30.17/0.8994/3.81          & 30.42/0.8650/3.51          & 31.19/0.8839/3.29          & 37.23/0.9343/2.09          \\
DPN \cite{kim2019deep}                            &                                 & 33.33/0.9059/2.57          & 31.07/0.8930/3.18           & 31.09/0.9019/3.40          & 31.91/0.8811/2.88          & 31.98/0.8840/2.96          & 37.49/0.9280/1.79          \\
\textbf{PIPNet}                    &                                 & \textbf{34.62/0.9353/2.14} & \textbf{32.13/0.9198/2.75} & \textbf{32.28/0.9285/2.89} & \textbf{32.88/0.9072/2.49} & \textbf{33.04/0.9140/2.50} & \textbf{39.44/0.9565/1.47} \\ \hline
\end{tabular}}
\caption{Quantitative evaluation of JDD on Quad Bayer CFA. A higher value of PSNR and SSIM indicates better results, while lower DeltaE indicates more colour consistency.}
\label{quadBayerTable}
\end{table*}

\begin{figure*}[!htb]%
\centering
\captionsetup[subfigure]{labelformat=empty}
\begin{minipage}{.28\textwidth}
    \begin{subfigure}{\textwidth}
      \centering
     \includegraphics[width=\textwidth,height=4.25cm]{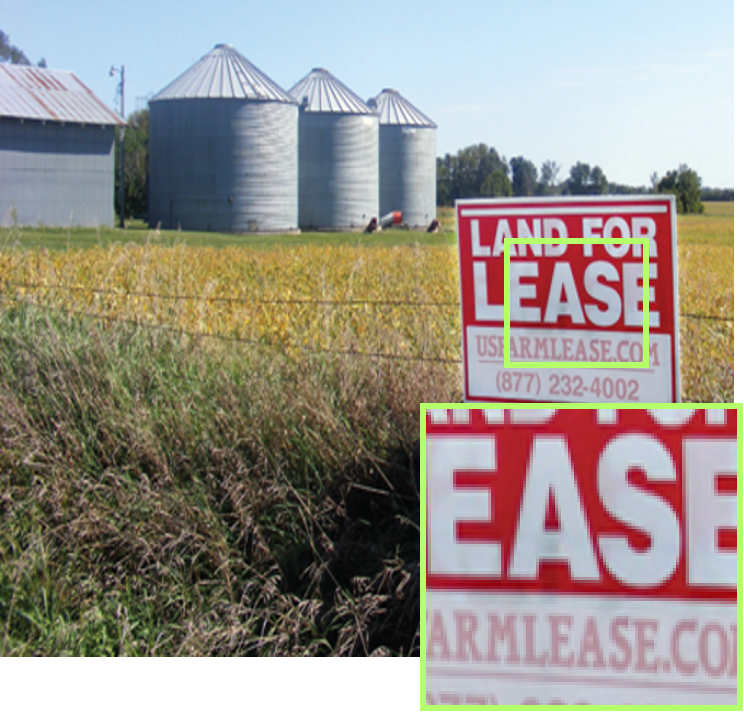}
     \caption{Reference}
    \end{subfigure}
  \end{minipage}
  \begin{minipage}{.13\textwidth}
    \begin{subfigure}{\textwidth}
      \centering
        \includegraphics[width=2.25cm,height=1.8cm]{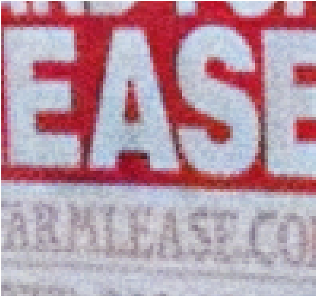}
        \caption{Deepjoint \cite{gharbi2016deep} }
    \end{subfigure}
    \begin{subfigure}{\textwidth}
      \centering
      \includegraphics[width=2.25cm,height=1.8cm]{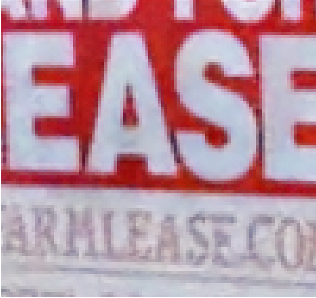}
      \caption{DeepISP \cite{schwartz2018deepisp}}
    \end{subfigure}
  \end{minipage}
  \begin{minipage}{.13\textwidth}
    \begin{subfigure}{\textwidth}
      \centering
        \includegraphics[width=2.25cm,height=1.8cm]{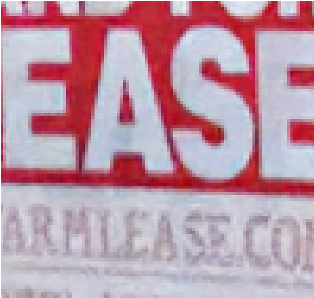}
        \caption{Kokkinos \cite{kokkinos2018deep} }
    \end{subfigure}
    \begin{subfigure}{\textwidth}
      \centering
      \includegraphics[width=2.25cm,height=1.8cm]{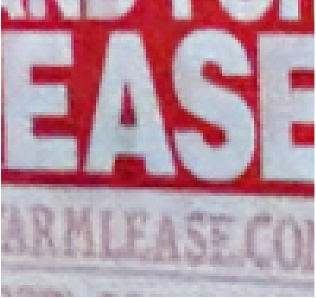}
      \caption{DPN \cite{kim2019deep} }
    \end{subfigure}
  \end{minipage}
  \begin{minipage}{.13\textwidth}
    \begin{subfigure}{\textwidth}
      \centering
      \includegraphics[width=2.25cm,height=1.8cm]{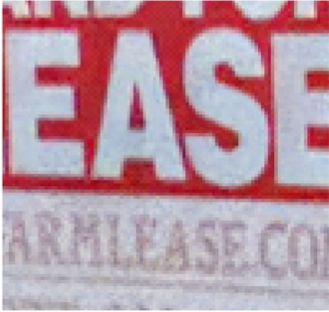}
      \caption{Dong \cite{dong2018joint}}
    \end{subfigure}
    \begin{subfigure}{\textwidth}
      \centering
     \includegraphics[width=2.25cm,height=1.8cm]{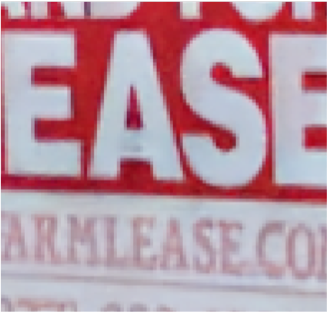}
     \caption{\textbf{PIPNet}}
    \end{subfigure}
  \end{minipage}
  \caption{ Qualitative evaluation of JDD on Quad Bayer CFA. }
\label{quadVis}
\end{figure*}

  Performing JDD on Quad Bayer CFA is substantially challenging. However, the proposed method aims to tackle this challenging task by using the novel PIPNet. Table. \ref{quadBayerTable} illustrates the performance comparison between the proposed PIPNet and target learning-based methods for Quad Bayer CFA. It is visible that our proposed method outperforms the existing learning-based methods in quantitative evaluation on benchmark datasets. Also, the visual results depicted in Fig. \ref{quadVis} confirm that the proposed method can reconstruct visually plausible images from Quad Bayer CFA.

%%%%%%%%%%%%%%%%%%%%%%%%%%%%%%%%%%%%%%%%%%
\subsubsection{Bayer CFA}

% \usepackage{multirow}
\begin{table*}[!htb]
\centering
\scalebox{.65}{\begin{tabular}{llllllll}
\hline
\multirow{2}{*}{\textbf{Model}} & \multirow{2}{*}{\textbf{$\sigma$}} & \textbf{BSD100}            & \textbf{WED}               & \textbf{Urban100}          & \textbf{McM}               & \textbf{Kodak}             & \textbf{linRGB}            \\ \cline{3-8}
                                &                                 & \textbf{PSNR/SSIM/DeltaE}  & \textbf{PSNR/SSIM/DeltaE}  & \textbf{PSNR/SSIM/DeltaE}  & \textbf{PSNR/SSIM/DeltaE}  & \textbf{PSNR/SSIM/DeltaE}  & \textbf{PSNR/SSIM/DeltaE}  \\ \hline
Deepjoint \cite{gharbi2016deep}                         & \multirow{6}{*}{5}              & 35.73/0.9661/2.21          & 32.96/0.9439/2.67          & 32.44/0.9468/2.99          & 33.56/0.9283/2.48          & 33.74/0.9519/2.62          & 38.99/0.9690/1.66           \\
Kokkinos \cite{kokkinos2018deep}                             &                                 & 36.81/0.9712/2.14          & 34.78/0.9561/2.43          & 34.30/0.9552/2.86          & 35.37/0.9462/2.20          & 35.24/0.9596/2.51          & 39.77/0.9716/1.56          \\
Dong \cite{dong2018joint}                           &                                 & 36.66/0.9710/2.29          & 33.72/0.9461/2.60          & 33.94/0.9486/2.85          & 33.88/0.9244/2.44          & 35.72/0.9646/2.56          & 38.07/0.9381/1.81          \\
DeepISP \cite{schwartz2018deepisp}                             &                                 & 38.60/0.9771/1.60          & 34.05/0.9497/2.38          & 33.53/0.9550/2.67          & 34.08/0.9327/2.31          & 35.29/0.9606/2.09          & 43.72/0.9826/1.11          \\
DPN \cite{kim2019deep}                           &                                 & 40.38/0.9837/1.25          & 36.71/0.9697/1.79          & 36.27/0.9707/2.08          & 37.21/0.9590/1.67          & 38.11/0.9739/1.61          & 43.97/0.9826/1.06          \\
\textbf{PIPNet}                   &                                 & \textbf{41.75/0.9851/1.11} & \textbf{37.84/0.9717/1.61} & \textbf{37.51/0.9731/1.87} & \textbf{38.13/0.9612/1.52} & \textbf{39.37/0.9768/1.41} & \textbf{46.41/0.9870/0.82} \\ \hline
Deepjoint \cite{gharbi2016deep}                         & \multirow{6}{*}{15}             & 33.50/0.9084/2.83          & 31.62/0.8988/3.15          & 31.20/0.9016/3.51          & 32.20/0.8829/2.92          & 31.99/0.8872/3.28          & 36.56/0.9201/2.15          \\
Kokkinos \cite{kokkinos2018deep}                             &                                 & 34.52/0.9377/2.62          & 32.93/0.9259/2.89          & 32.54/0.9264/3.30          & 33.52/0.9133/2.65          & 33.28/0.9202/2.98          & 37.82/0.9434/1.96          \\
Dong \cite{dong2018joint}                            &                                 & 33.86/0.9379/2.67          & 31.87/0.9123/2.95          & 32.20/0.9182/3.18          & 31.87/0.8792/2.80          & 33.27/0.9270/2.96          & 34.10/0.9009/2.28          \\
DeepISP \cite{schwartz2018deepisp}                            &                                 & 35.29/0.9410/2.23          & 32.46/0.9201/2.83          & 32.04/0.9260/3.12          & 32.57/0.9010/2.74          & 33.17/0.9192/2.68          & 39.81/0.9563/1.63          \\
DPN \cite{kim2019deep}                         &                                 & 36.90/0.9577/1.75          & 34.62/0.9462/2.22          & 34.33/0.9485/2.46          & 35.30/0.9355/2.04          & 35.28/0.9414/2.07          & 41.51/0.9726/1.36          \\
\textbf{PIPNet}                   &                                 & \textbf{37.59/0.9620/1.60} & \textbf{35.19/0.9494/2.07} & \textbf{34.93/0.9522/2.30} & \textbf{35.81/0.9391/1.92} & \textbf{35.96/0.9478/1.88} & \textbf{43.02/0.9735/1.13} \\ \hline
Deepjoint \cite{gharbi2016deep}                         & \multirow{6}{*}{25}             & 30.91/0.8240/3.82          & 29.73/0.8308/3.94          & 29.42/0.8345/4.33          & 30.25/0.8137/3.64          & 29.81/0.7963/4.28          & 34.09/0.8432/2.95          \\
Kokkinos \cite{kokkinos2018deep}                             &                                 & 32.20/0.8940/3.25          & 30.99/0.8864/3.47          & 30.64/0.8886/3.89          & 31.58/0.8717/3.18          & 31.19/0.8709/3.65          & 35.50/0.8987/2.47          \\
Dong \cite{dong2018joint}                           &                                 & 30.64/0.9005/3.52          & 29.39/0.8769/3.64          & 29.70/0.8888/3.88          & 29.34/0.8348/3.47          & 30.22/0.8866/3.90          & 30.61/0.8644/3.18          \\
DeepISP \cite{schwartz2018deepisp}                             &                                 & 32.54/0.8886/2.98          & 30.66/0.8765/3.41          & 30.33/0.8853/3.72          & 30.87/0.8547/3.28          & 31.03/0.8631/3.41          & 36.48/0.9063/2.31          \\
DPN \cite{kim2019deep}                           &                                 & 34.54/0.9267/2.21          & 32.70/0.9176/2.69          & 32.52/0.9237/2.89          & 33.42/0.9078/2.47          & 33.10/0.9044/2.53          & 39.20/0.9539/1.71          \\
\textbf{PIPNet}                   &                                 & \textbf{35.21/0.9379/2.01} & \textbf{33.24/0.9251/2.50} & \textbf{33.02/0.9300/2.71} & \textbf{33.95/0.9147/2.30} & \textbf{33.74/0.9186/2.31} & \textbf{40.46/0.9547/1.46} \\ \hline
\end{tabular}}
\caption{Quantitative evaluation of JDD on Bayer CFA. A higher value of PSNR and SSIM indicates better results, while lower DeltaE indicates more colour consistency.}
\label{bayerTable}
\end{table*}

\begin{figure*}%
\centering
\captionsetup[subfigure]{labelformat=empty}
\begin{minipage}{.28\textwidth}
    \begin{subfigure}{\textwidth}
      \centering
     \includegraphics[width=\textwidth,height=4.25cm]{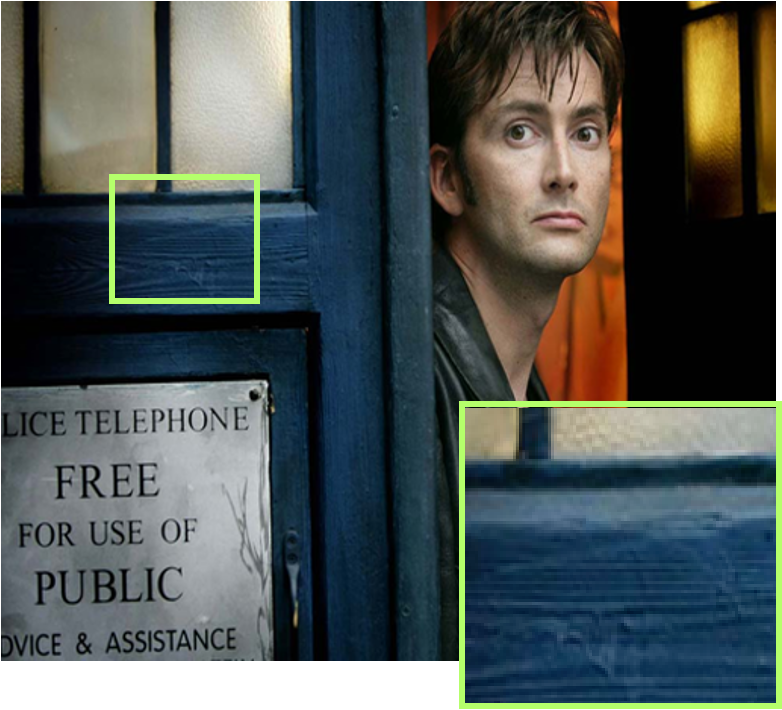}
     \caption{Reference}
    \end{subfigure}
  \end{minipage}
  \begin{minipage}{.13\textwidth}
    \begin{subfigure}{\textwidth}
      \centering
        \includegraphics[width=2.25cm,height=1.8cm]{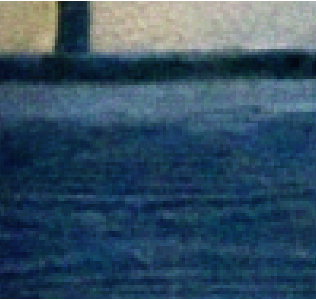}
        \caption{Deepjoint \cite{gharbi2016deep}}
    \end{subfigure}
    \begin{subfigure}{\textwidth}
      \centering
      \includegraphics[width=2.25cm,height=1.8cm]{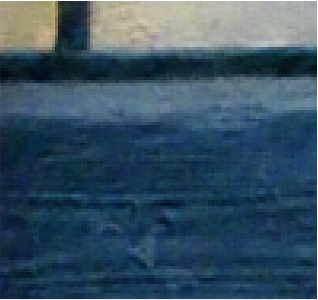}
      \caption{DeepISP \cite{schwartz2018deepisp}}
    \end{subfigure}
  \end{minipage}
  \begin{minipage}{.13\textwidth}
    \begin{subfigure}{\textwidth}
      \centering
        \includegraphics[width=2.25cm,height=1.8cm]{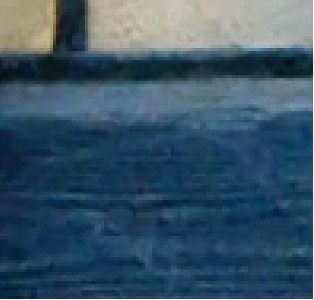}
        \caption{Kokkinos \cite{kokkinos2018deep} }
    \end{subfigure}
    \begin{subfigure}{\textwidth}
      \centering
      \includegraphics[width=2.25cm,height=1.8cm]{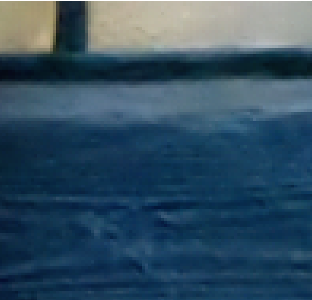}
      \caption{DPN \cite{kim2019deep} }
    \end{subfigure}
  \end{minipage}
  \begin{minipage}{.13\textwidth}
    \begin{subfigure}{\textwidth}
      \centering
      \includegraphics[width=2.25cm,height=1.80cm]{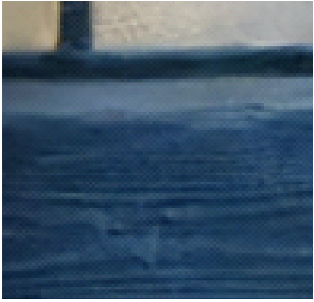}
      \caption{Dong \cite{dong2018joint}}
    \end{subfigure}
    \begin{subfigure}{\textwidth}
      \centering
     \includegraphics[width=2.25cm,height=1.80cm]{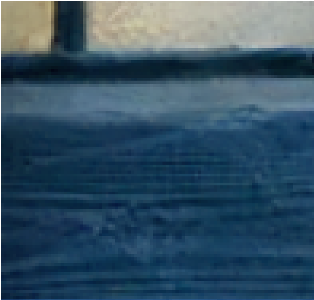}
     \caption{\textbf{PIPNet}}
    \end{subfigure}
  \end{minipage}
  \caption{ Qualitative evaluation of JDD on  Bayer CFA. }
  \label{bayer1}
\end{figure*}

As mentioned earlier, the pixel-bin image sensors have to employ Bayer CFA in numerous instances. Therefore, the proposed PIPNet has to perform JDD evenly on Bayer CFA. Table.  \ref{bayerTable} illustrates the JDD performance of the proposed method and its counterparts on Bayer CFA. The proposed method depicts the consistency in Bayer CFA as well. Also, it can recover more details while performing JDD on Bayer CFA without producing any visually disturbing artefacts, as shown in Fig. \ref{bayer1}.

%%%%%%%%%%%%%%%%%%%%%%%%%%%%%%%%%%%%%%%%%%

\begin{table}[!ht]
\scalebox{.7}{\begin{tabular}{lll}
\hline
\multicolumn{1}{c}{\multirow{2}{*}{\textbf{Model}}} & \textbf{sRGB}              & \textbf{linRGB}            \\ \cline{2-3}
\multicolumn{1}{c}{}                       & \textbf{PSNR/SSIM/DeltaE}  & \textbf{PSNR/SSIM/DeltaE}  \\ \hline
Base                                       & 24.51/0.7436/8.15          & 25.68/0.6258/8.71          \\
Base + AM                          & 32.82/0.9208/2.75          & 38.95/0.9430/1.77          \\
Base + AM + PCL                      & 33.96/0.9344/2.40           & 41.01/0.9647/1.31          \\
\textbf{Base + AM + PCL + RFL}          & \textbf{34.64/0.9436/2.22} & \textbf{41.82/0.9727/1.18} \\ \hline
\end{tabular}}
\caption{Ablation study on sRGB and linRGB images. Each component proposed throughout this study has an evident impact on network performance.}
\label{abliationTable}
\end{table}

\begin{figure}[!ht]
    \scalebox{.99}{\captionsetup[subfigure]{labelformat=empty,justification=centering}
     \centering
     \begin{subfigure}[b]{2cm}
         \centering
         \includegraphics[width=\textwidth]{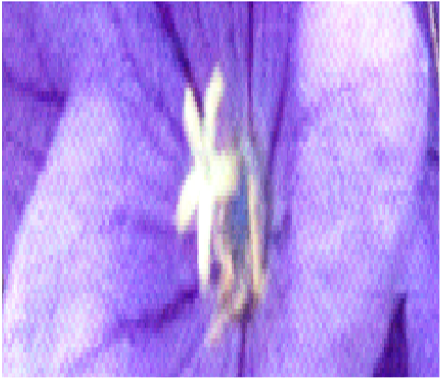}
         \caption{Base \newline}
         %\label{QBCFA}
     \end{subfigure}
     \hfill
     \begin{subfigure}[b]{2cm}
         \centering
         \includegraphics[width=\textwidth]{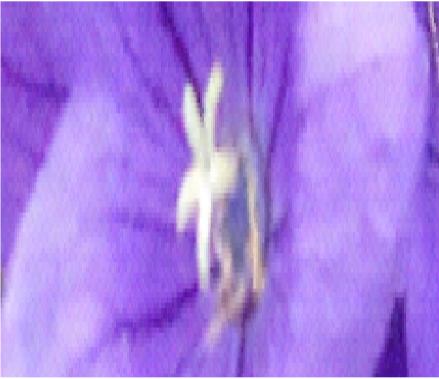}
         \caption{Base + AM\newline}
         %\label{BCFA}
     \end{subfigure}
     \hfill
     \begin{subfigure}[b]{2cm}
         \centering
         \includegraphics[width=\textwidth]{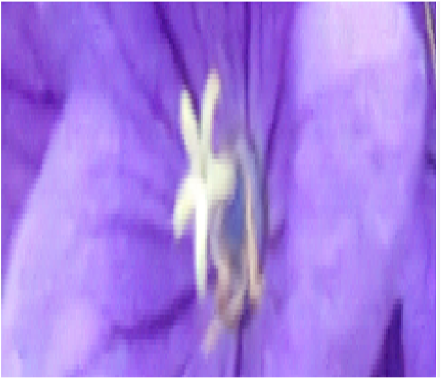}
         \caption{Base + AM\\ + PCL}
         %\label{QBCFA}
     \end{subfigure}
     \hfill
     \begin{subfigure}[b]{2cm}
         \centering
         \includegraphics[width=\textwidth]{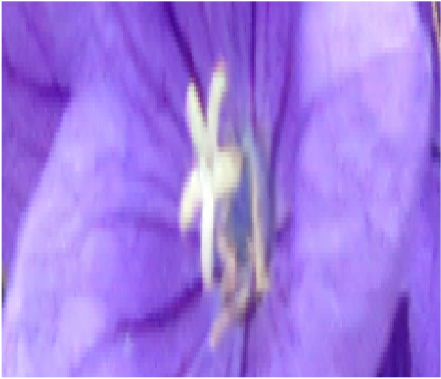}
         \caption{Base + AM \\+ PCL + RFL}
         %\label{BCFA}
     \end{subfigure}
     \hfill
      %\vspace{-1\baselineskip}
        \caption{Each proposed component plays a crucial role in JDD (best viewed in zoom). }%CA aims to reduce visual artefacts, PCL aims to improve colour consistency, and RFL aims to enhance details.}
        \label{ab-vis}}
\end{figure}

\subsection{Network analysis}
The practicability of the proposed network and its novel component has been verified by analyzing the network performance. 

\subsubsection{Ablation study}

An ablation study was conducted by removing all novel components like attention mechanism (AM), PCL, and RFL from the proposed method and later injecting them into the network consecutively. Fig. \ref{ab-vis} depicts the importance of each proposed component through visual results. Apart from that Table. \ref{abliationTable} confirms the practicability of novel components introduced by the proposed method. For simplicity, we combined all sRGB datasets and calculated the mean over the unified dataset while performing JDD on challenging Quad Bayer CFA. 

%%%%%%%%%%%%%%%%%%%%%%%%%%%%%%%%%%%%%%%%%%
\begin{table}[!ht]
\scalebox{.7}{\begin{tabular}{llll}
\hline
\multirow{2}{*}{\textbf{GD}} & \multirow{2}{*}{\textbf{Parameters}} & \textbf{sRGB}              & \textbf{linRGB}            \\ \cline{3-4}
                                        &                                      & \textbf{PSNR/SSIM/DeltaE}  & \textbf{PSNR/SSIM/DeltaE}  \\ \hline
1                                       & 2,778,863                            & 33.03/0.9239/2.73          & 38.30/0.9364/1.89          \\
2                                       & 3,118,703                            & 33.49/0.9308/2.61          & 39.79/0.9608/1.64          \\
\textbf{3}                              & \textbf{3,458,543}                   & \textbf{34.64/0.9436/2.22} & \textbf{41.82/0.9727/1.18} \\ \hline
\end{tabular}}
\caption{Group density vs model performance. The number of DAB blocks can impact network performance by making a trade-off between parameters and accuracy.}
\label{gdVsp}
\end{table}
\vspace{-1.\baselineskip}
\subsubsection{Group density vs performance}
\begin{figure}[!htb]
    \scalebox{.99}{\captionsetup[subfigure]{labelformat=empty,justification=centering}
     \centering
     \begin{subfigure}[b]{2cm}
         \centering
         \includegraphics[width=\textwidth]{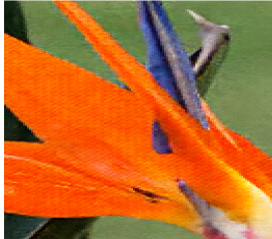}
         \caption{GD = 1}
         %\label{QBCFA}
     \end{subfigure}
     \hfill
     \begin{subfigure}[b]{2cm}
         \centering
         \includegraphics[width=\textwidth]{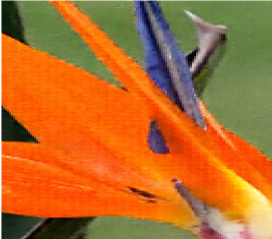}
         \caption{GD = 2}
         %\label{QBCFA}
     \end{subfigure}
     \hfill
     \begin{subfigure}[b]{2cm}
         \centering
         \includegraphics[width=\textwidth]{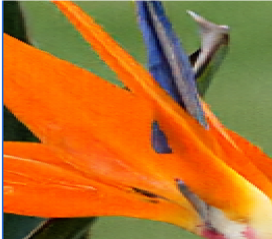}
         \caption{GD = 3}
         %\label{BCFA}
     \end{subfigure}
     \hfill
      %\vspace{-1\baselineskip}
        \caption{Impact of GD while performing JDD (best viewed in zoom).}
        \label{gd-vis}}
\end{figure}

Despite being significantly deeper and wider, the proposed PIPNet comprises 3.3 million parameters. The bottleneck block employed in GDAB allows our network to control the trainable parameter. Nevertheless, the number of parameters can be controlled by altering the group density (GD) of the GDAB blocks, as shown in Fig. \ref{gd-vis}. Additionally, Table. \ref{gdVsp} illustrates the relation between GD and performance in both colour spaces while performing JDD.

%%%%%%%%%%%%%%%%%%%%%%%%%%%%%%%%%%%%%%%%%%
\begin{figure*}[!htb]
    \captionsetup[subfigure]{labelformat=empty}
     \centering
     \begin{subfigure}[b]{\textwidth}
         \centering
         \includegraphics[width=13cm]{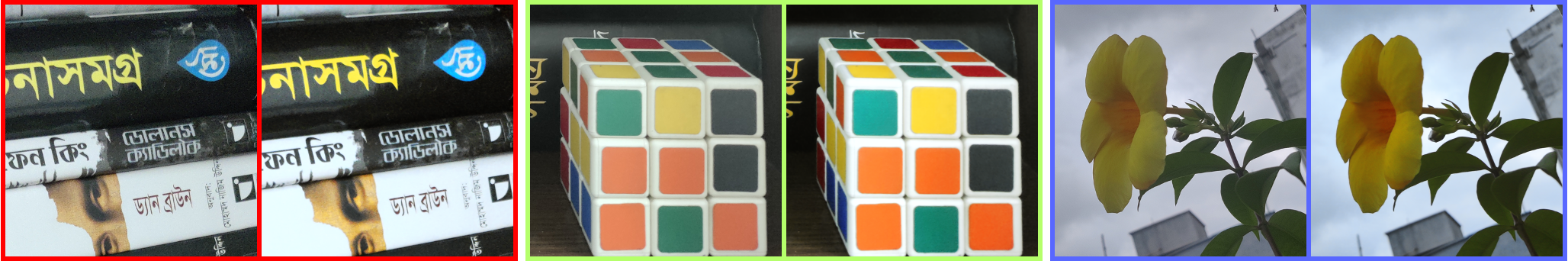}
         \caption{Quad Bayer reconstruction and enhancement}
         \label{QBISP}
     \end{subfigure}
     \hfill
     \begin{subfigure}[b]{\textwidth}
         \centering
         \includegraphics[width=13cm]{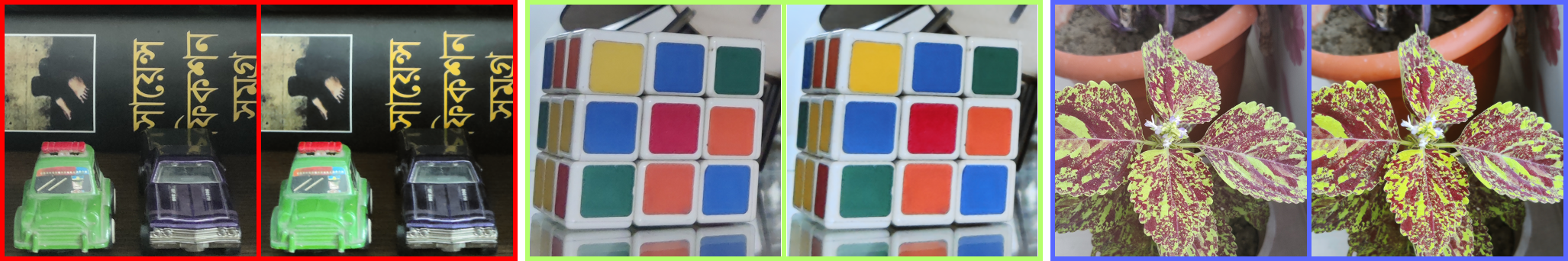}
         \caption{Bayer reconstruction and enhancement}
         \label{BISP}
     \end{subfigure}
      \vspace{-1\baselineskip}
        \caption{Qualitative comparison between pixel-bin image sensor output (i.e., Oneplus Nord) and the proposed PIPNet+. In every image pairs, Left: Onelpus Nord and Right: Results obtained by PIPNet+.  }
        \label{ISPRes}
\end{figure*}

\section{Image reconstruction and enhancement}
Typically, smartphone cameras are susceptible to produce flat, inaccurate colour profiles and noisy images comparing to professional cameras \cite{ignatov2017dslr, ignatov2020replacing}. To address this limitation and study the feasibility of a learning-based method on an actual pixel-bin image sensor (i.e., Sony IMX586), we stretch our PIPNet as a two-stage network. Stage-I of the extended network performs JDD, as described in section. \ref{experiment}, and stage-II aims to enhance reconstructed images' perceptual quality by correcting colour profile, white balancing, brightness correction, etc. The extended version of PIPNet is denoted as PIPNet+. It worth noting, the PIPNet+ comprises the same configuration (i.e., hyperparameters, GD, etc.) as its one stage variant; however, it has trained with smartphone-DSLR image pairs from the DPED dataset \cite{ignatov2017dslr}, as suggested in a recent study \cite{liang2019cameranet}. Our comprehensive solution's feasibility has compared with a recent smartphone (i.e., Oneplus Nord), which utilizes the pixel binning technique with actual hardware. We also develop an android application to control the binning process while capturing images for our network evaluation. Additionally, the captured images were resampled according to the CFA pattern prior to the model inference.

%%%%%%%%%%%%%%%%%%%%%%%%%%%%%%%%%%%%%%%%%%

\subsection{Visual results}
Fig. \ref{ISPRes} illustrates a visual comparison between Oneplus Nord and our PIPNet+. The proposed method can improve the perceptual quality of degraded images captured with an actual pixel-bin image sensor while performing ISP tasks like demosaicking, denoising, colour correction, brightness correction, etc.

%%%%%%%%%%%%%%%%%%%%%%%%%%%%%%%%%%%%%%%%%%
\subsection{User study}
Apart from the visual comparison, we perform a blind-fold user study comparing Oneplus Nord and our proposed method. Also, we develop a blind-fold online testing method, which allows the users to pick an image from pairs of Oneplus Nord and our reconstructed image. The testing evaluation process is hosted online publicly by an anonymous user. Thus, the unbiased user opinion can be cast to calculate the mean opinion score (MOS) for both CFA patterns. Table. \ref{mos} illustrates the MOS of our proposed method and Oneplus Nord. The proposed method outperforms Oneplus Nord in blind-fold testing by a substantial margin. Also, it confirms that Quad Bayer reconstruction is far more challenging than a typical Bayer reconstruction. Therefore, the traditional ISP illustrates deficiencies by producing visually pleasing images on such CFA patterns, while the proposed method can deliver more acceptable results.

\begin{table}[!htb]
\centering
\scalebox{.70}{\begin{tabular}{lll}
\hline
\textbf{CFA}                & \textbf{Method} & \textbf{MOS $\uparrow$} \\ \hline
\multirow{2}{*}{Quad Bayer} & Oneplus Nord            &       1.30       \\
                            & \textbf{PIPNet+}             &     \textbf{3.70}        \\ \hline
\multirow{2}{*}{Bayer}      & Oneplus Nord            &       1.50       \\
                            & \textbf{PIPNet+}            &      \textbf{3.50}       \\ \hline
\end{tabular}}
\caption{A user study on Oneplus Nord and PIPNet+. Higher MOS indicates better user preference.}
\label{mos}
\end{table}
%%%%%%%%%%%%%%%%%%%%%%%%%%%%%%%%%%%%%%%%%%

%%%%%%%%%%%%%%%%%%%%%%%%%%%%%%%%%%%%%%%%%%
\vspace{-1\baselineskip}
\section{Conclusion}
This study tackled the challenging task of performing JDD by incorporating a learning-based method specialized for a pixel-bin image sensor. We introduced a novel deep network that employed attention mechanisms and guided by a multi-term objective function, including two novel perceptual losses. Also, we stretched our proposed method to enhance the perceptual image quality of a pixel-bin image sensor along with reconstruction. The experiment with different CFA patterns illustrates that the proposed method can outperform the existing approaches in qualitative and quantitative comparison. Despite revealing new possibilities, we conducted our experiments mostly with simulated data collected by traditional Bayer sensors. Hence, the performance of the proposed network can differ in some complicated cases. It has planned to counter the data limitation by collecting a real-world dataset using pixel-bin image sensors for further study.

\section*{Acknowledgments}
This work was supported by the Sejong University Faculty Research Fund.

{\small
\bibliographystyle{ieee_fullname}

\begin{thebibliography}{10}\itemsep=-1pt

\bibitem{agranov2017pixel}
Gennadiy~A Agranov, Claus Molgaard, Ashirwad Bahukhandi, Chiajen Lee, and
  Xiangli Li.
\newblock Pixel binning in an image sensor, June~20 2017.
\newblock US Patent 9,686,485.

\bibitem{agustsson2017ntire}
Eirikur Agustsson and Radu Timofte.
\newblock Ntire 2017 challenge on single image super-resolution: Dataset and
  study.
\newblock In {\em IEEE Conf. Comput. Vis. Pattern Recog. Worksh.}, pages
  126--135, 2017.

\bibitem{aitken2017checkerboard}
Andrew Aitken, Christian Ledig, Lucas Theis, Jose Caballero, Zehan Wang, and
  Wenzhe Shi.
\newblock Checkerboard artifact free sub-pixel convolution: A note on sub-pixel
  convolution, resize convolution and convolution resize.
\newblock {\em arXiv preprint arXiv:1707.02937}, 2017.

\bibitem{barna2013method}
Sandor~L Barna, Scott~P Campbell, and Gennady Agranov.
\newblock Method and apparatus for improving low-light performance for small
  pixel image sensors, June~11 2013.
\newblock US Patent 8,462,220.

\bibitem{bayer1976color}
Bryce~E Bayer.
\newblock Color imaging array, July~20 1976.
\newblock US Patent 3,971,065.

\bibitem{chen2017sca}
Long Chen, Hanwang Zhang, Jun Xiao, Liqiang Nie, Jian Shao, Wei Liu, and
  Tat-Seng Chua.
\newblock Sca-cnn: Spatial and channel-wise attention in convolutional networks
  for image captioning.
\newblock In {\em IEEE Conf. Comput. Vis. Pattern Recog.}, pages 5659--5667,
  2017.

\bibitem{cordts2016cityscapes}
Marius Cordts, Mohamed Omran, Sebastian Ramos, Timo Rehfeld, Markus Enzweiler,
  Rodrigo Benenson, Uwe Franke, Stefan Roth, and Bernt Schiele.
\newblock The cityscapes dataset for semantic urban scene understanding.
\newblock In {\em IEEE Conf. Comput. Vis. Pattern Recog.}, pages 3213--3223,
  2016.

\bibitem{dai2019second}
Tao Dai, Jianrui Cai, Yongbing Zhang, Shu-Tao Xia, and Lei Zhang.
\newblock Second-order attention network for single image super-resolution.
\newblock In {\em IEEE Conf. Comput. Vis. Pattern Recog.}, pages 11065--11074,
  2019.

\bibitem{dong2018joint}
Weishong Dong, Ming Yuan, Xin Li, and Guangming Shi.
\newblock Joint demosaicing and denoising with perceptual optimization on a
  generative adversarial network.
\newblock {\em arXiv preprint arXiv:1802.04723}, 2018.

\bibitem{ehret2019joint}
Thibaud Ehret, Axel Davy, Pablo Arias, and Gabriele Facciolo.
\newblock Joint demosaicking and denoising by fine-tuning of bursts of raw
  images.
\newblock In {\em Int. Conf. Comput. Vis.}, pages 8868--8877, 2019.

\bibitem{fu2016fusion}
Xueyang Fu, Delu Zeng, Yue Huang, Yinghao Liao, Xinghao Ding, and John Paisley.
\newblock A fusion-based enhancing method for weakly illuminated images.
\newblock {\em Signal Process.}, 129:82--96, 2016.

\bibitem{gharbi2016deep}
Micha{\"e}l Gharbi, Gaurav Chaurasia, Sylvain Paris, and Fr{\'e}do Durand.
\newblock Deep joint demosaicking and denoising.
\newblock {\em ACM Trans. Graph.}, 35(6):1--12, 2016.

\bibitem{hirakawa2006joint}
Keigo Hirakawa and Thomas~W Parks.
\newblock Joint demosaicing and denoising.
\newblock {\em IEEE Trans. Image Process.}, 15(8):2146--2157, 2006.

\bibitem{hu2018squeeze}
Jie Hu, Li Shen, and Gang Sun.
\newblock Squeeze-and-excitation networks.
\newblock In {\em IEEE Conf. Comput. Vis. Pattern Recog.}, pages 7132--7141,
  2018.

\bibitem{ignatov2017dslr}
Andrey Ignatov, Nikolay Kobyshev, Radu Timofte, Kenneth Vanhoey, and Luc
  Van~Gool.
\newblock Dslr-quality photos on mobile devices with deep convolutional
  networks.
\newblock In {\em Int. Conf. Comput. Vis.}, pages 3277--3285, 2017.

\bibitem{ignatov2020replacing}
Andrey Ignatov, Luc Van~Gool, and Radu Timofte.
\newblock Replacing mobile camera isp with a single deep learning model.
\newblock In {\em IEEE Conf. Comput. Vis. Pattern Recog. Worksh.}, pages
  536--537, 2020.

\bibitem{khashabi2014joint}
Daniel Khashabi, Sebastian Nowozin, Jeremy Jancsary, and Andrew~W Fitzgibbon.
\newblock Joint demosaicing and denoising via learned nonparametric random
  fields.
\newblock {\em IEEE Trans. Image Process.}, 23(12):4968--4981, 2014.

\bibitem{kim2019deep}
Irina Kim, Seongwook Song, Soonkeun Chang, Sukhwan Lim, and Kai Guo.
\newblock Deep image demosaicing for submicron image sensors.
\newblock {\em J. Imaging Sci. Techn.}, 63(6):60410--1, 2019.

\bibitem{kim2019high}
Yongnam Kim and Yunkyung Kim.
\newblock High-sensitivity pixels with a quad-wrgb color filter and spatial
  deep-trench isolation.
\newblock {\em Sensors}, 19(21):4653, 2019.

\bibitem{kingma2014adam}
Diederik~P Kingma and Jimmy Ba.
\newblock Adam: A method for stochastic optimization.
\newblock {\em arXiv preprint arXiv:1412.6980}, 2014.

\bibitem{kokkinos2018deep}
Filippos Kokkinos and Stamatios Lefkimmiatis.
\newblock Deep image demosaicking using a cascade of convolutional residual
  denoising networks.
\newblock In {\em Eur. Conf. Comput. Vis.}, pages 303--319, 2018.

\bibitem{lahav2010color}
Assaf Lahav and David Cohen.
\newblock Color pattern and pixel level binning for aps image sensor using
  2$\times$ 2 photodiode sharing scheme, Aug.~10 2010.
\newblock US Patent 7,773,138.

\bibitem{lee2016automatic}
Joon-Young Lee, Kalyan Sunkavalli, Zhe Lin, Xiaohui Shen, and In~So Kweon.
\newblock Automatic content-aware color and tone stylization.
\newblock In {\em IEEE Conf. Comput. Vis. Pattern Recog.}, pages 2470--2478,
  2016.

\bibitem{liang2019cameranet}
Zhetong Liang, Jianrui Cai, Zisheng Cao, and Lei Zhang.
\newblock Cameranet: A two-stage framework for effective camera isp learning.
\newblock {\em arXiv preprint arXiv:1908.01481}, 2019.

\bibitem{liu2020joint}
Lin Liu, Xu Jia, Jianzhuang Liu, and Qi Tian.
\newblock Joint demosaicing and denoising with self guidance.
\newblock In {\em IEEE Conf. Comput. Vis. Pattern Recog.}, pages 2240--2249,
  2020.

\bibitem{liu2007high}
X Liu, Boyd Fowler, Hung Do, Steve Mims, Dan Laxson, and Brett Frymire.
\newblock High performance cmos image sensor for low light imaging.
\newblock In {\em International Image Sensor Workshop}, pages 327--330, 2007.

\bibitem{luo2001development}
M~Ronnier Luo, Guihua Cui, and Bryan Rigg.
\newblock The development of the cie 2000 colour-difference formula: Ciede2000.
\newblock {\em Color Research \& Application: Endorsed by Inter-Society Color
  Council, The Colour Group (Great Britain), Canadian Society for Color, Color
  Science Association of Japan, Dutch Society for the Study of Color, The
  Swedish Colour Centre Foundation, Colour Society of Australia, Centre
  Fran{\c{c}}ais de la Couleur}, 26(5):340--350, 2001.

\bibitem{ma2016waterloo}
Kede Ma, Zhengfang Duanmu, Qingbo Wu, Zhou Wang, Hongwei Yong, Hongliang Li,
  and Lei Zhang.
\newblock Waterloo exploration database: New challenges for image quality
  assessment models.
\newblock {\em IEEE Trans. Image Process.}, 26(2):1004--1016, 2016.

\bibitem{MartinFTM01}
D. Martin, C. Fowlkes, D. Tal, and J. Malik.
\newblock A database of human segmented natural images and its application to
  evaluating segmentation algorithms and measuring ecological statistics.
\newblock In {\em Int. Conf. Comput. Vis.}, volume~2, pages 416--423, July
  2001.

\bibitem{mechrez2018contextual}
Roey Mechrez, Itamar Talmi, and Lihi Zelnik-Manor.
\newblock The contextual loss for image transformation with non-aligned data.
\newblock In {\em Eur. Conf. Comput. Vis.}, pages 768--783, 2018.

\bibitem{mirza2014conditional}
Mehdi Mirza and Simon Osindero.
\newblock Conditional generative adversarial nets.
\newblock {\em arXiv preprint arXiv:1411.1784}, 2014.

\bibitem{pytorch}
Pytorch.
\newblock {PyTorch Framework} code.
\newblock \url{https://pytorch.org/}, 2016.
\newblock Accessed: 2020-11-14.

\bibitem{ronneberger2015u}
Olaf Ronneberger, Philipp Fischer, and Thomas Brox.
\newblock U-net: Convolutional networks for biomedical image segmentation.
\newblock In {\em International Conference on Medical image computing and
  computer-assisted intervention}, pages 234--241. Springer, 2015.

\bibitem{sandler2018mobilenetv2}
Mark Sandler, Andrew Howard, Menglong Zhu, Andrey Zhmoginov, and Liang-Chieh
  Chen.
\newblock Mobilenetv2: Inverted residuals and linear bottlenecks.
\newblock In {\em IEEE Conf. Comput. Vis. Pattern Recog.}, pages 4510--4520,
  2018.

\bibitem{schwartz2018deepisp}
Eli Schwartz, Raja Giryes, and Alex~M Bronstein.
\newblock Deepisp: Toward learning an end-to-end image processing pipeline.
\newblock {\em IEEE Trans. Image Process.}, 28(2):912--923, 2018.

\bibitem{strong2003edge}
David Strong and Tony Chan.
\newblock Edge-preserving and scale-dependent properties of total variation
  regularization.
\newblock {\em Inverse Probl.}, 19(6):S165, 2003.

\bibitem{tan2017joint}
Hanlin Tan, Xiangrong Zeng, Shiming Lai, Yu Liu, and Maojun Zhang.
\newblock Joint demosaicing and denoising of noisy bayer images with admm.
\newblock In {\em IEEE Int. Conf. Image Process.}, pages 2951--2955. IEEE,
  2017.

\bibitem{timofte2017ntire}
Radu Timofte, Eirikur Agustsson, Luc Van~Gool, Ming-Hsuan Yang, and Lei Zhang.
\newblock Ntire 2017 challenge on single image super-resolution: Methods and
  results.
\newblock In {\em IEEE Conf. Comput. Vis. Pattern Recog. Worksh.}, pages
  114--125, 2017.

\bibitem{wang2018esrgan}
Xintao Wang, Ke Yu, Shixiang Wu, Jinjin Gu, Yihao Liu, Chao Dong, Yu Qiao, and
  Chen Change~Loy.
\newblock Esrgan: Enhanced super-resolution generative adversarial networks.
\newblock In {\em Eur. Conf. Comput. Vis.}, pages 0--0, 2018.

\bibitem{woo2018cbam}
Sanghyun Woo, Jongchan Park, Joon-Young Lee, and In So~Kweon.
\newblock Cbam: Convolutional block attention module.
\newblock In {\em Eur. Conf. Comput. Vis.}, pages 3--19, 2018.

\bibitem{wu2011single}
Wei Wu, Zheng Liu, Wail Gueaieb, and Xiaohai He.
\newblock Single-image super-resolution based on markov random field and
  contourlet transform.
\newblock {\em J. Electron. Imaging}, 20(2):023005, 2011.

\bibitem{yanagawa2008kodak}
Akira Yanagawa, Alexander~C Loui, Jiebo Luo, Shih-Fu Chang, Dan Ellis, Wan
  Jiang, Lyndon Kennedy, and Keansub Lee.
\newblock Kodak consumer video benchmark data set: concept definition and
  annotation.
\newblock {\em Columbia University ADVENT Technical Report}, pages 246--2008,
  2008.

\bibitem{yoo2015low}
Yoonjong Yoo, Jaehyun Im, and Joonki Paik.
\newblock Low-light image enhancement using adaptive digital pixel binning.
\newblock {\em Sensors}, 15(7):14917--14931, 2015.

\bibitem{yuan2012automatic}
Lu Yuan and Jian Sun.
\newblock Automatic exposure correction of consumer photographs.
\newblock In {\em Eur. Conf. Comput. Vis.}, pages 771--785. Springer, 2012.

\bibitem{zhao2016loss}
Hang Zhao, Orazio Gallo, Iuri Frosio, and Jan Kautz.
\newblock Loss functions for image restoration with neural networks.
\newblock {\em IEEE Trans. Comput. Imag.}, 3(1):47--57, 2016.

\end{thebibliography}
}

\end{document}